\documentclass[12pt,letterpaper]{article}
\usepackage[top=1in,bottom=1in,left=1in,right=1in]{geometry}
\usepackage{amsmath,amssymb}
\usepackage{changepage}
\usepackage[utf8]{inputenc}
\usepackage{textcomp,marvosym}
\usepackage{cite}
\usepackage{nameref,hyperref}
\usepackage[right]{lineno}
\usepackage[table]{xcolor}
\usepackage{array}
\usepackage{apacite}
\usepackage{CJKutf8}
\usepackage{lastpage,fancyhdr,graphicx}
\usepackage{epstopdf}
\usepackage{tikz}
\usepackage{booktabs}
\usepackage{float}

\usepackage{soul} 
\soulregister\cite7
\soulregister\ref7
\soulregister\pageref7
\usepackage{xcolor}

\usepackage{listings}
\begin{document}
\vspace*{0.2in}

\begin{flushleft}
{\Large
\textbf{LLVMs4Protest: Harnessing the Power of Large Language and Vision Models for Deciphering Protests in the News}
}
\newline
\\
Yongjun Zhang\textsuperscript{1,2}
\\
\bigskip
\textbf{1} Department of Sociology and Institute for Advanced Computational Science, Stony Brook University, Stony Brook, New York, United States
\\
\textbf{2} Version 1, Updated on Nov 29, 2023
\\
\bigskip

* yongjun.zhang@stonybrook.edu

\end{flushleft}
\section*{Abstract}
Large language and vision models have transformed how social movements scholars identify protest and extract key protest attributes from multi-modal data such as texts, images, and videos. This article documents how we fine-tuned two large pretrained transformer models, including longformer and swin-transformer v2, to infer potential protests in news articles using textual and imagery data. First, the longformer model was fine-tuned using the Dynamic of Collective Action (DoCA) Corpus. We matched the New York Times articles with the DoCA database to obtain a training dataset for downstream tasks. Second, the swin-transformer v2 models was trained on UCLA-protest imagery data. UCLA-protest project contains labeled imagery data with information such as protest, violence, and sign. Both fine-tuned models will be available via \url{https://github.com/Joshzyj/llvms4protest}. We release this short technical report for social movement scholars who are interested in using LLVMs to infer protests in textual and imagery data.


\section*{Introduction}

The rapid advancement of large language and vision models has revolutionized the way social movement scholars analyze protest events using multimodal data such as text, images, and videos. Particularly, OpenAI's GPT-4 has empowered scholars with limited computational skills to intuitively examine textual and visual data. However, there are obstacles in adopting these AI tools. One is the cost of accessing advanced data analytic models from OpenAI, which can be prohibitive for analyzing vast quantities of data. An alternative is to use open-source LLMs like Meta's LLaMa 2, but this requires substantial machine learning knowledge and hardware resources. To address these challenges, we fine-tuned large-scale language and vision models for specific tasks, including protest event identification and key attribute extraction. These models are easily operable on standard computing devices.

Next, we will briefly summarize the major development in social movement literature that uses machine learning techniques to automatically identify protests. Then, we will delve into some technical details on training data and transfer learning to build our models. Finally, we will discuss some potential issues and limitations.

\section*{The Machine Learning Approach to Studying Social Movements}

Identifying protest events is one of the primary tasks in the field of social movement studies \cite{earl2004use}. One of the early endeavours made by social movement scholars was to systematically, manually extract protest information from newspaper articles. For instance, the Dynamics of Collective Action project identified protest events in 1955-1995 from the New York Times \cite{soule2005movement}. This human labeling strategy was both time consuming and expensive. Thus, many scholars attempted to automate the whole process using machine learning techniques. For instance, Alex Hanna developed the Machine-learning Protest Event Data System (MPEDS) to identify protest events and extract protest attributes from news articles \cite{hanna2017mpeds}. Other scholars such as Han Zhang and Jennifer Pan adopted Convolutional Neural Networks to classify multimodal social media data including texts and images to build Chinese protest database \cite{zhang2019casm}. In recent years, several groups in computer science have also devoted efforts to detect sociopolitical events on social media data. For instance, the shared tasks for Socio-political and Crisis Events Detection have attracted numerous social and computer scientists to develop deep learning algorithms to identify multilingual protests, fine-grained socio-political events, and Black Lives Matter protests in news \cite{hurriyetouglu2019overview}. Still, limited scholars have fine-tuned and shared large-scale language and vision models to identify protest events in news and social media data. One exception is that \citeA{davidson2023start} and \citeA{zhang2023generative} propose to use generative AI tools, specifically GPT-4V to extract protest information from images and texts.

Instead of relying on the recent advancement of GPT-4 models, in this article, we adopt the transfer learning approach to fine-tune existing pretrained large language and vision models for our downstream tasks in the field of social movement studies, specifically protest identification and attribute extraction. One of the underlying logic is that we attempt to take advantage of existing labeled training datasets both in textual and imagery format, which allows us to achieve the state-of-the-art performance.

\section*{Training Data Preparation}

\subsection*{Dynamics of Collective Action}
The widely used training dataset to classify protest events in newspaper articles is Dynamics of Collective Action \cite{earl2004use, hanna2017mpeds}. The DoCA project annotated news articles in the New York Times from 1960 to 1995. It is an event database, consisting of detailed article- and event-level information, such as reporting year, article title, protest size, claims, social movement organizations, locations, timing, etc (see \url{https://web.stanford.edu/group/collectiveaction/}). Unfortunately, DoCA does not contain the original articles for these events, thus we collected these textual data from ProQuest Database by searching article titles. We managed to obtain 16,700 articles that match the DoCA records. We further excluded articles containing multiple protest events in DoCA, which yielded 11,902 positive news articles in our training dataset.

Given that one of the primary tasks to identify whether a news article is protest related, we further retrieved negative articles from the New York Times Annotated Corpus via the Linguistic Data Consortium. We added 27,000 news articles in several categories that are not likely to contain protest related articles such as entertainment, book reviews, business, classified, finance, sports, real estate, leisure, and obituary.

\subsection*{UCLA-Protest Image Dataset}
We rely on the UCLA-Protest Image dataset to fine-tune our large vision model. UCLA protest database contains 40,764 images, including 11,659 protest images, with various annotations of visual attributes such as violence and police \cite{won2017protest}. We are also working on our own protest image dataset retrieving from Global Database of Events, Languages, and Tones (GDELT). 

\section*{Fine-tune Large Language and Vision Models}

\subsection*{The Longformer Model}

Instead of utilizing the pretrained BERT model, we fine-tune the longformer model, which is specifically used for the long documents \cite{beltagy2020longformer}. We fine-tune the longformer-base-4096 model with our training dataset, since it supports sequences of length up to 4096. Note that longformer-base-4096 is a BERT-like model started from the RoBERTa checkpoint and pretrained for marked language modeling on long documents. For more technical details, please check \citeA{beltagy2020longformer}'s paper.

\subsection*{The Swin-Transformer V2 Model}
Instead of training CNN models like VGG or ResNet, we fine-tune the Swin-transformer v2 model \cite{liu2021swin,liu2022swin}. Swin Transformer V2 has been shown the state-of-art performance in computer vision tasks. For more technical details, please check \citeA{liu2021swin,liu2022swin}'s articles. We fine-tuned the swinv2-base-patch4-window8-256 model pretrained by the Microsoft research group.

Note that the accuracy rate for both models is over 94\%. We will release more details on evaluation metrics when we finalize our models for scholarly community. The baseline models are available to our scholarly community.

\section*{Discussion}

\subsection*{AI for Social Movement Studies}

In this short essay, we attempt to document some of our ongoing work on fine-tuning large language and vision models to classify protest events and extract protest attributes from news articles and images.  In recent years, social activists, movement organizations, and bystanders are increasingly sharing protest activities via social media platforms with multimodal data and more visual information is included in news reports. Because of the recent advancement in generative AIs and deep learning techniques, scholars can take advantage of these large language and vision tools to analyze multimodal data to extract event information to address social science questions. Social movement scholars can benefit from adopting these tools to advance movement data collection and theory testing as well. Our work is one of these attempts by fine-tuning the state-of-the-art Longformer and Swin Transformer models to identify protests.

\subsection*{Limitations and Future Directions}
There are several limitations that readers should be aware of. First, fine-tuning transformer models requires scholars to have some basic knowledge of deep learning and programming skills. Fortunately, the swift advancement of generative AIs have lowered the barriers to computational social science \cite{zhang2023generative}. Scholars can use ChatGPT to generate, annotate, and debug codes. Second, the training data used to fine-tune longformer model has time constraints. As most of these positive news articles are from 1960-1995 and it might limit the potential of inferring protests in recent years. This warrants further studies. Third, our work still needs more cross-validation. For instance, a comparison between these fine-tuned models and GPT-4 annotation can offer social movement scholars more nuances in terms of choosing generative AIs or transfer learning with pretrained models. Lastly, our classifiers focus solely on protest identification and some limited number of protest attributes. We are still working on expanding protest attributes. These new features will be released soon once we finalize our fine-tuning processes.

\section*{Acknowledgments}
I acknowledge the support from the Institute for Advanced Computational Science for access to high performance computing systems and OpenAI APIs and ChatGPT. 

\bibliography{llvm4protest.bbl}

\begin{thebibliography}{}

\bibitem [\protect \citeauthoryear {%
Beltagy%
, Peters%
\BCBL {}\ \BBA {} Cohan%
}{%
Beltagy%
\ \protect \BOthers {.}}{%
{\protect \APACyear {2020}}%
}]{%
beltagy2020longformer}
\APACinsertmetastar {%
beltagy2020longformer}%
\begin{APACrefauthors}%
Beltagy, I.%
, Peters, M\BPBI E.%
\BCBL {}\ \BBA {} Cohan, A.%
\end{APACrefauthors}%
\unskip\
\newblock
\APACrefYearMonthDay{2020}{}{}.
\newblock
{\BBOQ}\APACrefatitle {Longformer: The long-document transformer} {Longformer: The long-document transformer}.{\BBCQ}
\newblock
\APACjournalVolNumPages{arXiv preprint arXiv:2004.05150}{}{}{}.
\PrintBackRefs{\CurrentBib}

\bibitem [\protect \citeauthoryear {%
Davidson%
}{%
Davidson%
}{%
{\protect \APACyear {2023}}%
}]{%
davidson2023start}
\APACinsertmetastar {%
davidson2023start}%
\begin{APACrefauthors}%
Davidson, T.%
\end{APACrefauthors}%
\unskip\
\newblock
\APACrefYearMonthDay{2023}{}{}.
\newblock
{\BBOQ}\APACrefatitle {Start Generating: Harnessing Generative Artificial Intelligence for Sociological Research} {Start generating: Harnessing generative artificial intelligence for sociological research}.{\BBCQ}
\newblock

\PrintBackRefs{\CurrentBib}

\bibitem [\protect \citeauthoryear {%
Earl%
, Martin%
, McCarthy%
\BCBL {}\ \BBA {} Soule%
}{%
Earl%
\ \protect \BOthers {.}}{%
{\protect \APACyear {2004}}%
}]{%
earl2004use}
\APACinsertmetastar {%
earl2004use}%
\begin{APACrefauthors}%
Earl, J.%
, Martin, A.%
, McCarthy, J\BPBI D.%
\BCBL {}\ \BBA {} Soule, S\BPBI A.%
\end{APACrefauthors}%
\unskip\
\newblock
\APACrefYearMonthDay{2004}{}{}.
\newblock
{\BBOQ}\APACrefatitle {The use of newspaper data in the study of collective action} {The use of newspaper data in the study of collective action}.{\BBCQ}
\newblock
\APACjournalVolNumPages{Annu. Rev. Sociol.}{30}{}{65--80}.
\PrintBackRefs{\CurrentBib}

\bibitem [\protect \citeauthoryear {%
Hanna%
}{%
Hanna%
}{%
{\protect \APACyear {2017}}%
}]{%
hanna2017mpeds}
\APACinsertmetastar {%
hanna2017mpeds}%
\begin{APACrefauthors}%
Hanna, A.%
\end{APACrefauthors}%
\unskip\
\newblock
\APACrefYearMonthDay{2017}{}{}.
\newblock
{\BBOQ}\APACrefatitle {Mpeds: Automating the generation of protest event data} {Mpeds: Automating the generation of protest event data}.{\BBCQ}
\newblock

\PrintBackRefs{\CurrentBib}

\bibitem [\protect \citeauthoryear {%
H{\"u}rriyeto{\u{g}}lu%
\ \protect \BOthers {.}}{%
H{\"u}rriyeto{\u{g}}lu%
\ \protect \BOthers {.}}{%
{\protect \APACyear {2019}}%
}]{%
hurriyetouglu2019overview}
\APACinsertmetastar {%
hurriyetouglu2019overview}%
\begin{APACrefauthors}%
H{\"u}rriyeto{\u{g}}lu, A.%
, Y{\"o}r{\"u}k, E.%
, Y{\"u}ret, D.%
, Yoltar, {\c{C}}.%
, G{\"u}rel, B.%
, Duru{\c{s}}an, F.%
\BDBL {}Akdemir, A.%
\end{APACrefauthors}%
\unskip\
\newblock
\APACrefYearMonthDay{2019}{}{}.
\newblock
{\BBOQ}\APACrefatitle {Overview of clef 2019 lab protestnews: Extracting protests from news in a cross-context setting} {Overview of clef 2019 lab protestnews: Extracting protests from news in a cross-context setting}.{\BBCQ}
\newblock
\BIn{} \APACrefbtitle {Experimental IR Meets Multilinguality, Multimodality, and Interaction: 10th International Conference of the CLEF Association, CLEF 2019, Lugano, Switzerland, September 9--12, 2019, Proceedings 10} {Experimental ir meets multilinguality, multimodality, and interaction: 10th international conference of the clef association, clef 2019, lugano, switzerland, september 9--12, 2019, proceedings 10}\ (\BPGS\ 425--432).
\PrintBackRefs{\CurrentBib}

\bibitem [\protect \citeauthoryear {%
Liu%
\ \protect \BOthers {.}}{%
Liu%
\ \protect \BOthers {.}}{%
{\protect \APACyear {2022}}%
}]{%
liu2022swin}
\APACinsertmetastar {%
liu2022swin}%
\begin{APACrefauthors}%
Liu, Z.%
, Hu, H.%
, Lin, Y.%
, Yao, Z.%
, Xie, Z.%
, Wei, Y.%
\BDBL {}others%
\end{APACrefauthors}%
\unskip\
\newblock
\APACrefYearMonthDay{2022}{}{}.
\newblock
{\BBOQ}\APACrefatitle {Swin transformer v2: Scaling up capacity and resolution} {Swin transformer v2: Scaling up capacity and resolution}.{\BBCQ}
\newblock
\BIn{} \APACrefbtitle {Proceedings of the IEEE/CVF conference on computer vision and pattern recognition} {Proceedings of the ieee/cvf conference on computer vision and pattern recognition}\ (\BPGS\ 12009--12019).
\PrintBackRefs{\CurrentBib}

\bibitem [\protect \citeauthoryear {%
Liu%
\ \protect \BOthers {.}}{%
Liu%
\ \protect \BOthers {.}}{%
{\protect \APACyear {2021}}%
}]{%
liu2021swin}
\APACinsertmetastar {%
liu2021swin}%
\begin{APACrefauthors}%
Liu, Z.%
, Lin, Y.%
, Cao, Y.%
, Hu, H.%
, Wei, Y.%
, Zhang, Z.%
\BDBL {}Guo, B.%
\end{APACrefauthors}%
\unskip\
\newblock
\APACrefYearMonthDay{2021}{}{}.
\newblock
{\BBOQ}\APACrefatitle {Swin transformer: Hierarchical vision transformer using shifted windows} {Swin transformer: Hierarchical vision transformer using shifted windows}.{\BBCQ}
\newblock
\BIn{} \APACrefbtitle {Proceedings of the IEEE/CVF international conference on computer vision} {Proceedings of the ieee/cvf international conference on computer vision}\ (\BPGS\ 10012--10022).
\PrintBackRefs{\CurrentBib}

\bibitem [\protect \citeauthoryear {%
Soule%
\ \BBA {} Earl%
}{%
Soule%
\ \BBA {} Earl%
}{%
{\protect \APACyear {2005}}%
}]{%
soule2005movement}
\APACinsertmetastar {%
soule2005movement}%
\begin{APACrefauthors}%
Soule, S.%
\BCBT {}\ \BBA {} Earl, J.%
\end{APACrefauthors}%
\unskip\
\newblock
\APACrefYearMonthDay{2005}{}{}.
\newblock
{\BBOQ}\APACrefatitle {A movement society evaluated: Collective protest in the United States, 1960-1986} {A movement society evaluated: Collective protest in the united states, 1960-1986}.{\BBCQ}
\newblock
\APACjournalVolNumPages{Mobilization: An International Quarterly}{10}{3}{345--364}.
\PrintBackRefs{\CurrentBib}

\bibitem [\protect \citeauthoryear {%
Won%
, Steinert-Threlkeld%
\BCBL {}\ \BBA {} Joo%
}{%
Won%
\ \protect \BOthers {.}}{%
{\protect \APACyear {2017}}%
}]{%
won2017protest}
\APACinsertmetastar {%
won2017protest}%
\begin{APACrefauthors}%
Won, D.%
, Steinert-Threlkeld, Z\BPBI C.%
\BCBL {}\ \BBA {} Joo, J.%
\end{APACrefauthors}%
\unskip\
\newblock
\APACrefYearMonthDay{2017}{}{}.
\newblock
{\BBOQ}\APACrefatitle {Protest activity detection and perceived violence estimation from social media images} {Protest activity detection and perceived violence estimation from social media images}.{\BBCQ}
\newblock
\BIn{} \APACrefbtitle {Proceedings of the 25th ACM international conference on Multimedia} {Proceedings of the 25th acm international conference on multimedia}\ (\BPGS\ 786--794).
\PrintBackRefs{\CurrentBib}

\bibitem [\protect \citeauthoryear {%
H.~Zhang%
\ \BBA {} Pan%
}{%
H.~Zhang%
\ \BBA {} Pan%
}{%
{\protect \APACyear {2019}}%
}]{%
zhang2019casm}
\APACinsertmetastar {%
zhang2019casm}%
\begin{APACrefauthors}%
Zhang, H.%
\BCBT {}\ \BBA {} Pan, J.%
\end{APACrefauthors}%
\unskip\
\newblock
\APACrefYearMonthDay{2019}{}{}.
\newblock
{\BBOQ}\APACrefatitle {Casm: A deep-learning approach for identifying collective action events with text and image data from social media} {Casm: A deep-learning approach for identifying collective action events with text and image data from social media}.{\BBCQ}
\newblock
\APACjournalVolNumPages{Sociological Methodology}{49}{1}{1--57}.
\PrintBackRefs{\CurrentBib}

\bibitem [\protect \citeauthoryear {%
Y.~Zhang%
}{%
Y.~Zhang%
}{%
{\protect \APACyear {2023}}%
}]{%
zhang2023generative}
\APACinsertmetastar {%
zhang2023generative}%
\begin{APACrefauthors}%
Zhang, Y.%
\end{APACrefauthors}%
\unskip\
\newblock
\APACrefYearMonthDay{2023}{}{}.
\newblock
{\BBOQ}\APACrefatitle {Generative AI has lowered the barriers to computational social sciences} {Generative ai has lowered the barriers to computational social sciences}.{\BBCQ}
\newblock
\APACjournalVolNumPages{arXiv preprint arXiv:2311.10833}{}{}{}.
\PrintBackRefs{\CurrentBib}

\end{thebibliography}

\end{document}